\documentclass{llncs}
\usepackage{graphicx}
\usepackage{amsmath, amssymb}
\usepackage[mathscr]{euscript}
\usepackage{tikz}
\usepackage{multirow} 
\usepackage[para]{footmisc}
\usepackage[misc]{ifsym}

\makeatletter
\newcommand{\thanksymbol}[1]{%
  \textsuperscript{\@fnsymbol{#1}}%
}
\makeatother

\title{\framework{}: A Framework for\\Safe Deep Reinforcement Learning for Autonomous Driving}
\author{Jaeyoung Lee\thanks{contributed equally}
\and Aravind Balakrishnan\thanksymbol{1}
\and Ashish Gaurav\thanksymbol{1}
\and\\ Krzysztof Czarnecki\,\textsuperscript{\Letter}
\and Sean Sedwards\thanksymbol{1}
}
\institute{University of Waterloo, Canada}

\newcommand{\framework}{\textsc{WiseMove}}
\newcommand{\col}[1]{\ensuremath{\{#1\}^\intercal}}
\newcommand{\hlinew}[1]{%
\noalign{\global\arrayrulewidth#1}%
\hline%
\noalign{\global\arrayrulewidth0.4pt}
}
\newcommand{\true}{{\sf true}}
\newcommand{\false}{{\sf false}}

\begin{document}
\maketitle
\begin{abstract}
Machine learning can provide efficient solutions to the complex problems encountered in autonomous driving, but ensuring their safety remains a challenge.
A number of authors have attempted to address this issue, but there are few publicly-available tools to adequately explore the trade-offs between functionality, scalability, and safety.

We thus present \framework, a software framework to investigate safe deep reinforcement learning in the context of motion planning for autonomous driving.
\framework{} adopts a modular learning architecture that suits our current research questions and can be adapted to new technologies and new questions.
We present the details of \framework, demonstrate its use on a common traffic scenario, and describe how we use it in our ongoing safe learning research.
\end{abstract}

\section{Introduction}\label{sec:introduction}
Learned systems seem essential in autonomous driving, with machine learning providing performance that is difficult to replicate using other approaches.
The safety of learned systems is, however, difficult to guarantee, creating a conflict of requirements.

Autonomous driving requires sophisticated perception and decision making, which is often most efficiently achieved by machine learning.
The state space of the resulting complex functions is typically intractable to analysis, especially when considering the hybrid dynamics of autonomous driving.
Safety is nevertheless essential, so a number of authors have already attempted to address the notion of safe learning in this context (e.g.~\cite{AD14,AFGKZT14,SSS16,PRHGK17}).
Many practical challenges remain, however, and we are not aware of any suitable software tools to investigate the complex trade-offs between safety and functionality.

Our group\footnote{uwaterloo.ca/waterloo-intelligent-systems-engineering-lab/} has developed an autonomous driving software stack that has already been used to drive autonomously for 100 km~\cite{autonomoose100k},
so our interest in safety is more than just academic.
Our pressing concern is the safety and scalability of motion planning, where safety can refer to both the learning and deployment phases, and scalability refers to both the creation and verification of a motion planning solution.

Our current stack employs deep neural networks for perception, but the rest of the motion planning architecture is more conventionally implemented. Governed by an overall mission planner, high-level decisions, such as {\em slow down}, {\em turn left}, etc., are made by a ``behaviour planner'' that resolves hand-coded logical constraints (rules) over a discrete abstraction of the environment ({\em approaching the stop region}, {\em in the intersection}, {\em stopped}, etc.). To implement the chosen high-level behaviour, a low-level reference trajectory is generated by a ``local planner'' that performs an optimization with respect to jerk and rate of progress, etc., over classes of smooth trajectories that avoid obstacles and pass through the current continuous state and future waypoints. A controller on the vehicle then actuates the reference trajectory.

The recent success of deep reinforcement learning (DRL) in playing Go~\cite{AlphaGo,AlphaGoZero}, and its success with other applications having intractable state space~\cite{Lillicrap2015}, suggests DRL as a scalable way to implement motion planning for autonomous vehicles.
We have thus devised a hierarchical and modular DRL framework, \framework{}, to investigate the trade-offs between safety, performance and scalability.
A DRL-based approach using a similar architecture to our own has recently been proposed in~\cite{PRHGK17}.
The work reports interesting results, but provides no software tool or other means to verify them.
In what follows, having described \framework{}, we present results of experiments that can be reproduced by installing our publicly-available code\footnote{\label{foot:repo}git.uwaterloo.ca/wise-lab/wise-move}.

\begin{figure} [t]
\begin{minipage}[t]{0.48\textwidth}
   \centering
    \begin{tikzpicture}
    \scriptsize
	\draw[every node/.style={draw,shape=rectangle,minimum width=0.5\textwidth}]
    	(0,2.4) node [draw,minimum width=0.6\textwidth,minimum height=11mm] (env) {}
    	(0,2.65) node [draw=none]{Environment}
    	(0,2.2) node {Vehicle Models}
	(0,0.05) node[draw,minimum width=0.6\textwidth,minimum height=30mm](mcts){}
	(0,1.3) node[draw=none] {MCTS}
	(0,0.9) node (hlp){High-level Policy}
	(0,-0.1) node [minimum height=10mm](opt){}
	(0,0.15) node [draw=none] {Options}
	(0,-1.1) node (ver){LTL Verification};
	\draw[every node/.style={draw,fill=black!20,minimum size=4mm}]
    	(-1,-0.3) node {}
	(-0.5,-0.3) node {}
	(0,-0.3) node {}
	(0.5,-0.3) node {}
	(1,-0.3) node {};
	\draw[->,thick] (env.east) to [bend left]node[above,rotate=-90]{state}(mcts.east);
	\draw[->,thick] (mcts.west) to [bend left]node[above,rotate=90]{option}(env.west);
	\draw[->,thick] (hlp)->(opt);
	\draw[->,thick](opt)--(ver);
    \end{tikzpicture}
    \caption{Diagrammatic representation of \framework{} planning architecture.}
    \label{fig:planning-architecture}
\end{minipage}
~
\begin{minipage}[t]{0.48\textwidth}
	\includegraphics[width=\textwidth]{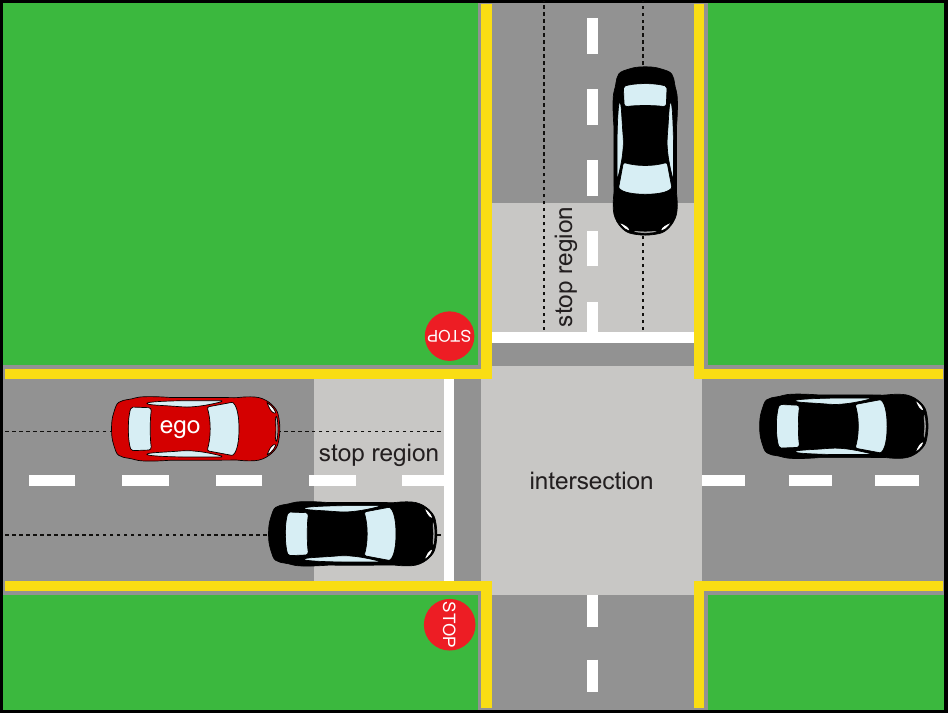}
    \caption{Visualization of the simple intersection environment.}
    \label{fig:road geometry}
\end{minipage}
\end{figure}

\subsubsection{\bfseries\framework{}} is an options-based modular safe DRL framework, written in Python.
Its hierarchical structure is designed to mirror the architecture of our existing software stack, learning (approximately) safe high- and low-level decision-making policies that are then made safer using Monte Carlo tree search (MCTS~\cite{kocsis2006bandit,chaslot2008monte}).
Options~\cite{Sutton1999} model primitive manoeuvres, to which are associated low-level policies that implement them.
These policies are learned separately, in advance, each using a deep neural network to encode the continuous action space.
A learned high-level policy over options decides which option to take in any given situation.
This high-level policy corresponds to the behaviour planner in our existing software stack; the low-level policies correspond to our ``local'' planner and controller.
To define safe behaviour and option termination conditions, \framework{} uses runtime verification to validate simulation traces and assign rewards during both learning and planning with MCTS.
MCTS is an expected-outcome algorithm~\cite{Abramson1990}, which \framework{} uses to perform a stochastic look-ahead, to choose the safest next option.

Fig.~\ref{fig:planning-architecture} gives a diagrammatic overview of \framework{}'s planning architecture.
The current state is provided to the planner (MCTS) by the environment.
MCTS explores and verifies hypothesized future trajectories, using the learned high-level policy over options as a baseline.
MCTS chooses the best next option it discovers, which is then used to update the environment.

\section{Dynamics}
\label{sec:description of the road scenario}

In this section we describe the dynamics of \framework{}, as used in the simple intersection scenario shown in Fig.~\ref{fig:road geometry}.
We first introduce the discrete and continuous dynamics, then use these to construct a partially-observable Markov decision process (POMDP) representing the full dynamics of the (simplified) driving task we consider in our experiments.

\subsubsection{Continuous Dynamics.}
Let $x_{i, \mathsf{veh}} := (X_i, Y_i, \theta_i, \,v_i, \,\psi_i)$ and $u_i := (\mathfrak{a}_i, \, \rho_i)$ be the continuous state and control input, respectively, of vehicle $i$, with position $(X_i,Y_i)$, speed $v_i$, acceleration $\mathfrak{a}_i$, heading angle $\theta_i$, steering angle $\psi_i$, rate of change of steering angle $\rho_i$, and wheel base $L$.
Then its continuous dynamics are 
\begin{equation}
	\begin{cases}
    \begin{aligned}
		&{\dot X}_i = v_i \sin \theta_i && {\dot Y}_i = v_i \cos \theta_i 
        \qquad \qquad {\dot \theta}_i = v_i \tan (\psi_i/L) \quad
        \\[5pt]
		&{\dot v}_i \,= \mathfrak{a}_{i} \quad (|\mathfrak{a}_{i}| \leq \mathfrak{a}_\mathrm{max}) && 
         {\dot \psi}_i = \rho_{i} \quad (|\rho_{i}| \leq \rho_\mathrm{max}, \; |\psi_i| \leq \psi_\mathrm{max}).
	\end{aligned}
    \end{cases}
   \hspace{-1.5em}
    \label{eq:vehicle dynamics}
\end{equation}
The state is updated every $\Delta t$ (by numerical integration) according to state transition function ${x}_{i, \mathsf{veh}}' = g_{\mathsf{veh}}(x_{i, \mathsf{veh}}, u_i)$.
We use the previous input, $u_{i, \mathsf{prev}} := (\mathfrak{a}_{i, \mathsf{prev}}, \, {\rho}_{i, \mathsf{prev}})$, to approximate the jerk by ${\dot{\mathfrak{a}}}_i \approx (\mathfrak{a}_{i} - \mathfrak{a}_{i, \mathsf{prev}}) / \Delta t$. Defining $x_i := (x_{i, \mathsf{veh}}, u_{i,\mathsf{prev}})$, the complete continuous dynamics of vehicle $i$ is $x_i' = g_\mathsf{c}(x_i, u_i)$,
where continuous transition function $g_\mathsf{c}(x_i, u_i) := (g_\mathsf{veh}(x_{i, \mathsf{veh}}, u_i), u_{i})$. 

\subsubsection{Discrete Dynamics.} Vehicle $i$ has discrete state
\[
    z_i := (\mathtt{has\_stopped\_in\_stop\_region}_i,\mathtt{has\_entered\_stop\_region}_i,\mathit{waited}_i),
\]
where $waited_i\in\mathbb{Z}$ is initialised to $-1$, incremented every $\Delta t$ the vehicle remains in the stop region, and reset to $-1$ when it leaves.
The discrete state is updated according to discrete transition function $z_i' = g_\mathsf{d}(z_i, y_{i,\mathsf{loc}}')$, where $y_{i,\mathsf{loc}}$ is a local discrete abstraction of the continuous state $x_i$.
The global hybrid state is given by~$s := (\mathbf{x}, \mathbf{z})\in\mathscr{S}$, with $\mathbf{x} := \col{x_i}$ and $\mathbf{z} := \col{z_i}$, and where $\col{\cdot}$ denotes the column vector formed by all the indexed elements.
The global discrete output of vehicle $i$ is denoted $y_{i, \mathsf{glob}}$, comprising $s$ and $\col{y_{i,\mathsf{loc}}}$.
The complete discrete output of vehicle $i$ is  $y_i := (y_{i,\mathsf{loc}}, y_{i, \mathsf{glob}})$, and $\mathbf{y} = \col{y_i}$ denotes the global discrete output.

\subsubsection{Complete POMDP.}
The full update dynamics is given by
\begin{equation}
s'=(\mathbf{x}',\mathbf{z}')^\intercal = (f_\mathsf{c}(s, a),f_\mathsf{d}(s, a))^\intercal := f(s, a),\label{eq:full hybrid dynamics}
\end{equation}
where action~$a\in\mathscr{A}=[-\mathfrak{a}_\mathrm{max}, \mathfrak{a}_\mathrm{max}] \times [- \rho_\mathrm{max}, \rho_\mathrm{max}]$ is the ego vehicle's control input~$u_0$, $f_\mathsf{c}(s,a) := \col{g_\mathsf{c}(x_i, u_i)}$, and $f_\mathsf{d}(s,a) := \col{g_\mathsf{d}(z_i, h_\mathsf{loc} \big (g_\mathsf{c}(x_i, u_i))\big )}$.
Typically, we adopt a different policy, $\mu$, for the non-ego vehicles. E.g., the aggressive driving policy of~\cite{PRHGK17}.
Hence, $u_i = \mu(s)$ for all $i \neq 0$.

Omitting the index from any variable of the ego vehicle, the ego's full observation~$o \in \mathscr{O}$ is 
\begin{equation}
o := (\mathbf{x}, \mathbf{z}, \mathbf{y}, \alpha_1, \alpha_2, \cdots, \alpha_N),\label{eq:observation}
\end{equation}
with $\alpha_j := (X - X_j, Y - Y_j, v_j, \mathfrak{a}_{j,\mathsf{prev}}, \mathit{waited}_j)$.
The dynamics~\eqref{eq:full hybrid dynamics} and observation~\eqref{eq:observation} induce a POMDP,
$\mathscr{P} := (\mathscr{S}, \mathscr{S}^0, \mathscr{A}, f, \mathscr{O}, h)$,
where $\mathscr{S}^0 \subseteq \mathscr{S}$ is the set of initial states and $h(s) := (\mathbf{x},\mathbf{z},\mathbf{y}, \alpha_1, \alpha_2, \cdots, \alpha_N)$ is the observation function. As~\eqref{eq:full hybrid dynamics} and~\eqref{eq:observation} are deterministic, the only stochasticity, if any, is the randomly-chosen control action during training and/or planning.

\section{Features and Architecture}\label{sec:software features and architecture}

\framework{} comprises four high-level Python modules:
\texttt{verifier}, the incremental verifier module;
 \texttt{env}, the environments module;
 \texttt{options}, the options module;
 and \texttt{backends}, the DRL module.

\subsection{Incremental Verification}
\label{sec:incremental model checking}
The \texttt{verifier} module provides methods for checking temporal logic property strings that are constructed according to the following LTL-like syntax:
\begin{equation}
\varphi = \texttt{F}\;\varphi\mid\texttt{G}\;\varphi\mid\texttt{X}\;\varphi\mid\varphi\;\texttt{=>}\;\varphi\mid\varphi\;\texttt{or}\;\varphi\mid\varphi\;\texttt{and}\;\varphi\mid\texttt{not}\;\varphi\mid\varphi\;\texttt{U}\;\varphi\mid\texttt{(}\varphi\texttt{)}\mid\alpha\label{eq:logic}
\end{equation}
Literal symbols \texttt{U}, \texttt{F}, \texttt{G} and \texttt{X} are the standard {\em until}, {\em eventually} ({\em finally}), {\em always} ({\em globally}) and {\em next-state} temporal operators, respectively.
The other literal symbols have their obvious meanings. 
Atomic propositions, $\alpha$, evaluate to \true{} or \false{} in each state and are represented by human readable strings.
This syntax is compatible with the properties used by other logic-based safe learning approaches and is sufficiently expressive for our current needs.
In what follows we refer to LTL properties, meaning properties written according to~\eqref{eq:logic}.

The verifier works as an efficient, incremental ``learntime'' or runtime monitor, constructed using the Coco/R compiler generator~\cite{coco-r}. 
Efficiency of the verifier is important, since properties are checked in the innermost loops of learning the low-level and high-level policies, as well as within the MCTS.

In \framework, an episode continues while its termination conditions are all \false{}.
If one of the conditions becomes \true{}, the episode terminates with a positive or negative terminal reward, depending on whether the termination represents success or failure, respectively.
\framework{} also expresses traffic rules using temporal logic, e.g.,
\begin{align*}
    \varphi_1 &= \text{\texttt{G}(\texttt{in\_stop\_region} \texttt{=>} (\texttt{in\_stop\_region} \texttt{U}  \texttt{has\_stopped\_in\_stop\_region}))}, \\
    \varphi_2 &= \text{\texttt{G}(\texttt{in\_intersection} \texttt{=>} \texttt{intersection\_is\_clear})}, \textrm{and} \\
    \varphi_3 &= \text{\texttt{G}(\texttt{not\;in\_intersection} \texttt{U} \texttt{highest\_priority})}.
\end{align*}
The result of verification at each time step determines whether to terminate the episode and, if so, the appropriate terminal reward.
LTL expressions are also used to represent, if any, the preconditions and goal conditions of each option.
Some options and preconditions are listed in Table~\ref{tab:manouevres}.

\subsection{Environments and Backends}

The \texttt{env} module provides support for environments that adhere to the OpenAI Gym~\cite{openai-gym} interface, meaning they implement \texttt{step}, \texttt{reset} and \texttt{render} functions, which can respectively update, initialize and visualize the environment. The \texttt{options} module also adheres to this interface. This standardization allows for a plug-and-play functionality, such that we can plug in any gym-compliannt environment.

The \texttt{backends} module provides the architecture for control logic specification. Control logic for the environment, whether \emph{learned} through optimization or \emph{programmed} imperatively, have pre-defined abstract interfaces in \texttt{backends} that can be implemented as desired.

For the purpose of our experiments, we use DDPG~\cite{Lillicrap2015}, DQN~\cite{mnih2013playing} and MCTS for the different levels of our hierarchical architecture, which are all connected through the \texttt{options} module. The \texttt{options} module provides three modes of high level logic specification: \texttt{rl}, \texttt{mcts} and \texttt{manual}. While \texttt{rl} and \texttt{mcts} respectively use DQN and MCTS for high level logic, the \texttt{manual} mode provides support for a deterministic options graph, the transitions of which are manually defined. Our implementation currently use keras\footnote{http://keras.io} and  keras-rl\footnote{http://github.com/keras-rl/keras-rl} for DRL training, and the training hierarchy can be specified through a \texttt{json} file. The configuration of the hierarchy can also be altered (via the \texttt{json} file) to use other options.

\subsection{Options and Learning}\label{sec:options}

The \texttt{options} module provides the hierarchical decision-making architecture for RL.
An option is an elementary manoeuvre, hence we use the notation $m \in \mathscr{M}$ to denote elements of the set of options, and define $m := (\mathscr{S}_m^0, \pi_m, \beta_m)$ \cite{Sutton1999}.
$\mathscr{S}_m^0 \subseteq \mathscr{S}$ is the initiation set in which $m$ is available, $\pi_m:\mathscr{F} \to \mathscr{A}$ is a low-level policy---a map from the feature space $\mathscr{F}\ni\phi$ to the action space~$\mathscr{A}\ni a$---and $\beta_m$ is the termination condition, which can evaluate to \true{} or \false{}. 

In \framework, we learn the low-level policy $\pi_m$ for each option $m$ first, and then learn the high-level policy~$\Pi: \mathscr{F} \to \mathscr{M}$, which determines the option $m = \Pi(\phi)$ at each decision instant.
Only options whose preconditions are satisfied may be chosen. Table~\ref{tab:manouevres} shows examples of the logical preconditions.

If an option~$m$ is chosen by the high-level decision maker, i.e., $\Pi$ or MCTS, at discrete time $t$, the agent generates a sequence of actions $a_{t} \, a_{t+\Delta{t}} \, a_{t+2\Delta{t}} \cdots$ according to $a_\tau = \pi_m(\phi_\tau)$, for all $\tau \geq t$ until termination, and where $\phi_\tau$ is the feature vector at time $\tau$. 
An option $m$ terminates when the termination condition $\beta_m$ becomes \true{}. In this case, the agent chooses the next option and executes it until it terminates. This process continues until the whole episode ends. 

The options module defines the termination condition~$\beta_m$ of each option as the disjunction of (i) a violation of an LTL requirement, (ii) successful completion, (iii) collision, and (iv) timeout.

\setlength{\tabcolsep}{3.5pt}
\renewcommand{\arraystretch}{1.4}

	\begin{table} [t] \centering
		\begin{tabular} {|c|*2{l|}}
		    \hline
			   Option & Description &  Example LTL Precondition  \\
			   \hlinew{1pt}
			   \texttt{KeepLane} & keep lane while driving & -  \\
			    \hline 
			   \texttt{Stop} & stop at the stop region & \texttt{G(not\;has\_stopped\_in\_stop\_region)}  \\
			   \hline 
			   \texttt{Wait} & 
			   \begin{tabular}{@{}l@{}}			   
			   	wait at the stop region \\[-5pt]
			   	then drive forward
			   \end{tabular}
				& 
			   \begin{tabular}{@{}l@{}}			   
			   \texttt{G((has\_stopped\_in\_stop\_region and} \\[-5pt]
			   \hspace{2em}\texttt{in\_stop\_region) U highest\_priority)}
			   \end{tabular}  \\
			   \hline			   
			   \texttt{Follow} &  follow vehicle ahead & \texttt{G(veh\_ahead)} \\
			   \hline
			   \texttt{ChangeLane} & change to other lane & \texttt{G(not(in\_intersection\;or\;in\_stop\_region)\!)} \\
			   \hline 
		\end{tabular}
                  \vspace{10pt}
		\caption{The options and examples of their preconditions.}
		\label{tab:manouevres}
	\end{table}
		
\subsubsection{Learning} 

The objective of DRL is to learn a set of low-level policies, $\{\pi_m^* : \mathscr{F} \to \mathscr{A} \}_{m \in \mathscr{M}}$, and a high-level policy, $\Pi^*: \mathscr{F} \to \mathscr{M}$, that jointly maximize the value function
\begin{equation}
    V(s_{0:T}, a_{0:T}, m_{0:K}) := - \sum_{t = 0}^{T - 1} \gamma^t \!\! \cdot \mathsf{inst}(o_t, a_t, m_{k}) + \gamma^T \!\! \cdot \mathsf{term}(o_T),
    \label{eq:cumulative reward}
\end{equation}
where:
$\mathsf{inst}(\cdot)$ is the instantaneous reward;
$\mathsf{term}(\cdot)$ is the terminal reward;
$s_0 \in \mathscr{S}^0$ is the initial state;
$\gamma \in (0, 1)$ is the discount rate;
$T, K \in \mathbb{N} \cup \{ \infty \}$ are the terminal time and decision instants of an episode, respectively;
$s_t$ is the state at time $t$;
$o_t$ is the ego's observation at time $t$;
$m_{k} = \Pi(\phi_{t_k})$ is the option chosen by the high-level policy at decision time $t_k$ of decision instant $k$, applied until its termination at the next decision time, $t_{k+1}$;
and $a_t = \pi_{m_k}(\phi_t)$ is the action given by the low-level policy at times $t = t_k, t_k + \Delta t, t_k + 2\Delta t, \cdots, t_{k+1} - \Delta t$.

Instantaneous reward $\mathsf{inst}(o_t, a_t, m_k)$ is given at every time $t$, after observing~$o_t$ and taking action~$a_t$ under the option~$m_k$.
It is calculated as the weighted Euclidean norm of continuous features, such as the speed error (the difference between the reference speed and the actual speed) and the lateral position error (the distance from the lane centre line).

The terminal reward, $\mathsf{term}(o_T)$, is given following the terminal observation, $o_T$, at terminal time $T$.
In practice, we give a large positive reward for successfully reaching the goal of the option, or a large negative reward in the case of a collision or violation of an LTL requirement.

\section{Experiments}\label{sec:experiments}

We report here the results of experiments performed on the intersection environment illustrated in Fig.~\ref{fig:road geometry}.
The road scenario contains the ego vehicle, initially placed on the horizontal route, and up to $6$ other vehicles placed at random on either the horizontal or vertical routes.
Vehicles are placed in either the left or right lane at random; they drive to the right on the horizontal route and drive down on the vertical route.
Vehicles must stop completely in the stop region.
No left or right turn is allowed, but vehicles can change lane within the same route.
With this configuration, the goal of the ego vehicle is to arrive at the right end of the route without any collision or violation of the traffic rules, while respecting the speed limit.
An episode terminates when any collision or violation occurs with the ego vehicle (failure), or if it reaches the goal region with its longitudinal speed less than or equal to the speed limit (success).

We first trained the low-level option-specific policies using DDPG, under various constraints.
These include those that express preconditions (as given in Table~\ref{tab:manouevres}), as well as additional properties used to enhance the training of low-level policies.
The additional constraints include liveness conditions (e.g., \texttt{G(not stopped\_now)}) for promoting exploration, and safety-related properties (e.g., \texttt{G(not veh\_ahead\_too\_close)} in \texttt{Follow}).
Table~\ref{tab:low level w and w/o LTLs} shows the significant performance gains achieved when using the additional properties.

\begin{table} [b] 
    \centering
	\begin{tabular}{|c||*5{c|}}
	    \hline
 Add'l LTL & \texttt{KeepLane} & \texttt{Stop} & \texttt{Wait}& \texttt{Follow} & \texttt{ChangeLane} \\
        \hlinew{1pt}
 Unused & 15.8 (25.7) & 45.7 (30.2) & 0.00 (0.00) & 58.2 (19.6) & 52.2 (35.7) \\
        \hline 
 Used & 75.2 (37.9) & 94.0 (10.6) & 93.8 (6.12) & 78.0 (17.9) & 93.7 (15.7) \\
        \hline
	\end{tabular}
         \vspace{10pt}
	\caption{Performance of low-level policies trained for $10^5$ steps, with and without additional LTL properties: mean (std) of the number of success in 100 episodes, averaged over 10 trials with independent retraining for each.}  
	\label{tab:low level w and w/o LTLs}
\end{table}

We next trained the low-level policy for each option using $10^6$ steps, in order to have a set of high performance base manoeuvres.
We then trained high-level policies using DQN.
The results in Table~\ref{tab:high and MCTS} show that without MCTS, the high-level policy achieves an average success rate of 89.7\%, with a small number of collisions and traffic rule violations.
With MCTS, the average success rate jumps to 98.0\%, with even fewer violations and collisions.

\begin{table} [t]
	\centering
	\begin{tabular}{|*3{c|}|*3{c|}}
		\hline
		\multicolumn{3}{|c||}{Without MCTS} & \multicolumn{3}{c|}{With MCTS} \\
		\hlinew{1pt}
		 success & violation & collision & success & violation & collision \\
		\hline		
		89.7 (2.66) & 4.24 (1.88) & 4.41 (1.81) & 98.0 (1.05) & 0.60 (0.51)  & 1.40 (0.96) \\
		\hline
	\end{tabular}
         \vspace{10pt}
	\caption{Performance of the high-level decision-making, without and with MCTS: mean (std) \% of success/violation/collision rates in the experiment, averaged over 10 trials with 100 episodes. Without MCTS, each high-level policy for each trial is independently retrained for $2\times10^5$ steps. With MCTS, all trials used a high-level policy that achieved 92\% success. Both experiments were conducted with low-level policies pre-trained with $10^6$ steps, using additional properties to enhance training.}
	\label{tab:high and MCTS}
\end{table}

The standard deviations of our results indicate that they are robust and repeatable.
The full details of our experimental setups, along with scripts to reproduce the results, can be found in our repository\textsuperscript{\ref{foot:repo}}.

\section{Conclusion and Prospects}\label{sec:conclusion}

Our early results with \framework{} have been successful in reproducing published results and revealing interesting phenomena and challenges that are not apparent in the literature.
For example, the unpredictable interactions between different constraints and rewards. 
Having revealed these, however, we are confident that \framework{} is well equipped to help explore them.

Our ongoing research will use  \framework{} with different scenarios and more complex vehicle dynamics.
We will also exploit \framework's modularity, using different types of non-ego vehicles (aggressive, passive, learned, programmed, etc.) and interleave learned components with programmed components from our autonomous driving software stack.

\subsection*{Acknowledgment}
This work is partly supported by the Japanese Science and Technology agency (JST) ERATO project JPMJER1603: HASUO Metamathematics for Systems Design.

\bibliographystyle{splncs04}
\bibliography{main}
\end{document}